\documentclass[10pt,twocolumn,letterpaper]{article}

\usepackage[pagenumbers]{wacv} 

\usepackage{graphicx}
\usepackage{amsmath}
\usepackage{amssymb}
\usepackage{booktabs}
\usepackage{enumitem}
\usepackage{soul}

\usepackage[pagebackref,breaklinks,colorlinks]{hyperref}

\usepackage[capitalize]{cleveref}
\crefname{section}{Sec.}{Secs.}
\Crefname{section}{Section}{Sections}
\Crefname{table}{Table}{Tables}
\crefname{table}{Tab.}{Tabs.}
\usepackage{tabularx}
\usepackage{multirow}
\usepackage{booktabs}

\usepackage{upquote}

\begin{document}

\title{Tackling Data Bias in MUSIC-AVQA: Crafting a Balanced Dataset for Unbiased Question-Answering}

\author{
Xiulong Liu \thanks{These authors contributed equally.} \\
University of Washington\\
{\tt \small xl1995@uw.edu}
\and
Zhikang Dong \footnotemark[1] \\
Stony Brook University\\
{\tt \small zhikang.dong.1@stonybrook.edu}
\and
Peng Zhang \thanks{Corresponding author}\\
Bytedance.com\\
{\tt \small zhang.peng@bytedance.com}
}

\maketitle

\begin{abstract}
In recent years, there has been a growing emphasis on the intersection of audio, vision, and text modalities, driving forward the advancements in multimodal research.
However, strong bias that exists in any modality can lead to the model neglecting the others. Consequently, the model's ability to effectively reason across these diverse modalities is compromised, impeding further advancement.

In this paper, we meticulously review each question type from the original dataset, selecting those with pronounced answer biases. To counter these biases, we gather complementary videos and questions, ensuring that no answers have outstanding skewed distribution. In particular, for binary questions, we strive to ensure that both answers are almost uniformly spread within each question category. As a result, we construct a new dataset, named MUSIC-AVQA v2.0, which is more challenging and we believe could better foster the progress of AVQA task. Furthermore, we present a novel baseline model that delves deeper into the audio-visual-text interrelation. On MUSIC-AVQA v2.0, this model surpasses all the existing benchmarks, improving accuracy by 2\% on MUSIC-AVQA v2.0, setting a new state-of-the-art performance.
\end{abstract}

\section{Introduction}
\label{sec:intro}
\begin{figure*}[t]
    \centering
  \includegraphics[width = 0.9\linewidth]{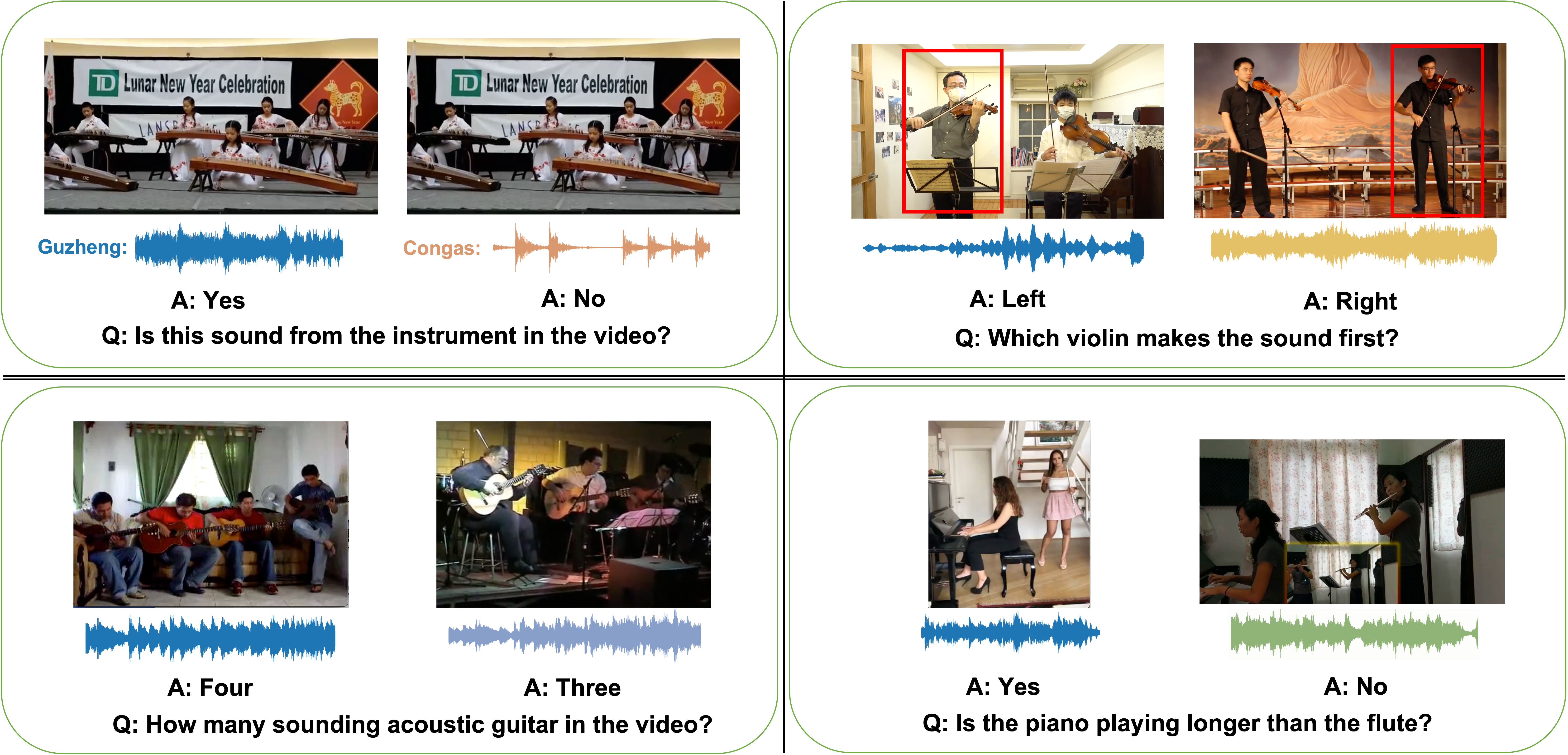}
  \caption{An overview of \textit{MUSIC-AVQA v2.0 QA} samples. (a) showcases two videos with the same visuals but different audio; (b) both videos display two violinists in different orders; (c) videos present acoustic guitar ensembles of different counts; (d) features a piano-flute duet where the piano plays longer in the first video. To answer accurately, models must consider these audio-visual nuances rather than just language priors.}
  \label{fig:task}
\vspace{-6mm}
\end{figure*}
In audio-visual learning, the interplay between audio and visual information provides a rich avenue for understanding dynamic scenarios. One particular task that embodies this synergy is Audio-Visual Question Answering (AVQA).
As Fig.~\ref{fig:task} shows,
given a musical instrument performance video and associated music audio, models are expected to answer questions that are related to them
or the relationships therein.
Unlike existing popular Visual Question Answering (VQA) task, which only tackles two modalities - vision and language, AVQA is designed to bridge and reason through all three modalities - vision, language and audio, which stimulates the merge of a new dataset. MUSIC-AVQA dataset, proposed to cater for this task, acting as an important benchmark, facilitate the research progress in this field.

The dataset consists of 3 major question categories by modality: Audio-Visual, Visual and Audio questions, respectively. Across all the categories, 5 question aspects are considered, including ``Existential" - e.g. Is there a voiceover?, ``Temporal" - e.g. Which violin makes the sound first?, ``Counting" - e.g. How many sounding violins in the video?, ``location" - e.g. Where is the performance?, and "Comparative" - e.g. Which object makes the sound first?. By joining the modality types with question aspects, the dataset ends up with 33 question templates, in which one can change the instrument name from one to another based on the video. Each question template contains a fix set of answers, ranging from binary answers (e.g. "yes" and "no") to counting answers (e.g. "1", "2", "3" etc.) and so on. Finally, MUSIC-AVQA dataset contains 9,290 (7,423 real, 1,867 synthetic) videos and 45,867 corresponding questions.

We notice there is a strong bias exists in the dataset, which results in undermining the reliability of this dataset as a credible benchmark. For example, in a particular question category asking about whether or not the audio track comes from the instrument played in the video, Over $90\%$ of answer in the dataset is "yes". In audio-visual temporal question, when asking which instrument in the video sounds first, nearly 80\% answer is ``Simultaneously''. In counting questions, answers of small number like 1 and 2 alone could take up more than 50\%. These imbalance exhibit across question aspects of ``Existence", ``Counting", ``Temporal", ``Location" and ``Comparative''. Such outstanding bias could impact the model training negatively and thus lowering the ability of the underlying model, as model will be trained to favor towards to the most common answers in the training set and ignoring the importance of video and audio, as well as the reasoning between three of them. This problem has been explored in other tasks, for example, Visual Question Answwering (VQA)~\cite{agrawal2016vqa} task, on which several
datasets and models~\cite{zhang2016yin,agrawal2018dont,cadene2020rubi,kv2020reducing} are proposed to tackle it. But in AVQA, this problem is still remain untouched. In light of these observations, we endeavor to systematically address the issue and propose an improved version of the data, called MUSIC-AVQA v2.0.

Our initial step was to evaluate the answer distribution for each question template. Through this analysis, we identified specific templates that displayed a skewed answer distribution. After identifying these skewed templates, we proceed to select the question templates that have minority answers, designating them as our primary targets for balance. Recognizing the significance of holistic data representation, we take the additional step of manually collecting videos that corresponded with these specific question template and answer pairs. This ensures that the dataset is more representative and balanced.

Existing works have proposed models to the task of AVQA, which were trained on the public MUSIC-AVQA dataset. On the evaluation of the balanced dataset, we show that the strong data bias does misleading the model training, resulting in inferior performance. Moreover, we also contribute another model that has the ability to learning the connections across all the three different modalities. The model extends existing methods by: (i). adding an additional pretrained Audio-Spectrogram-Transformer (AST) branch for audio-visual grounding, and (ii). designing a cross-modal pixel-wise attention between audio and visual spatial maps. Experiment results on balanced dataset indicate that our new baseline, even without audio-visual pretraining, surpasses the performance of existing state-of-the-art models by more than 2\%.
In summary, the primary contributions of this work are 3-folds:

(i). We identify and mitigate the data imbalance issue in existing MUSIC-AVQA dataset, and establish MUSIC-AVQA v2.0, a more balanced AVQA benchmark;
(ii). MUSIC-AVQA v2.0 enriches the MUSIC-AVQA dataset with an additional 1204 real videos compared to exising 7.4k real videos, leading to an additional inclusion of 8.1k QA pairs. A substantial portion of the additional videos are musical instrument ensemble involving 3 or more instruments, capturing more complex audio-visual relationship;
(iii). We introduce a novel baseline model, which demonstrates enhanced capability in bridging different modalities through the use of pretrained models and an attention mechanism. Upon evaluation on both bias and balanced datasets, our model surpassed all existing baselines, establishing a new and stronger baseline.

\section{Related Works}
Emerging as a significant branch of multi-modal learning, audio-visual learning explores the complex relationship between audio and visual signals. Various tasks have been established under this domain, such as audio-visual source separation~\cite{zhao2018sound,gao2018learning,zhao2019sound,gan2020music,Gao_2019_ICCV,tian2021cyclic,chatterjee2021visual}, audio-visual event localization~\cite{tian2018audio,Wu_2019_ICCV,xuan2020cross,Duan_2021_WACV,Xia_2022_CVPR}, audio-visual generation (audio-to-visual or visual-to-audio)~\cite{shlizerman2018audio,ginosar2019learning,lee2019dancing,su2020audeo,FoleyMusic2020,su2020multi,su2021does,di2021video}, audio-visual detection and segmentation.

More recently, audio-visual question answering~\cite{yun2021panoavqa,yang2022avqa,li2022learning} emerges as a novel task as a tri-modal extension of visual (video) question answering (VQA). Such a QA task focuses on the aspect of understanding the video by combining both audio and visual context. Pano-AVQA~\cite{yun2021panoavqa} set up a QA benchmark that focused on 360 degree videos. Later, MUSIC-AVQA~\cite{li2022learning} proposed a new benchmark that focused on musical instrument performance videos, and the questions cover more aspects of the audio-visual associations by incorporating challenging questions like counting, temporal relation and comparative questions. Both MUSIC-AVQA and Pano-AVQA designed QA benchmark to be open-ended QA. In contrast, AVQA~\cite{yang2022avqa} provided a multiple-choice format, adding depth to the question types by including causal and purpose questions.

In addition to these tasks, different representation learning paradigms or model architecture designs have been proposed to effectively bridge the gap between audio and visual modality. In representation learning aspect, early works exploit the audio-visual correspondence~\cite{NEURIPS2020_328e5d4c,Morgado_2021_CVPR} by designing tasks like audio-visual matching and audio-visual synchronization. More recently, a self-supervised learning-based framework~\cite{gong2023contrastive} that combines both contrastive learning~\cite{chen2020simple,chen2020improved,grill2020bootstrap} and masked auto-encoder technique~\cite{he2021masked} was proposed to pre-train audio-visual representations. In model architecture aspect, LAVISH~\cite{lin2023vision} designed a lightweight adapter layer that bridges between vision pre-trained backbones. During training, only the adapter layer along with the task-specific layers are trained while the remaining parameters are kept frozen, which enables parameter-efficient learning without large-scale pre-training. The method achieves the state-of-art performance on 3 different tasks with benchmarks, respectively audio-visual event localization (AVE~\cite{tian2018audiovisual}), audio-visual segmentation (AVS~\cite{zhou2023audiovisual}) and audio-visual question answering (MUSIC-AVQA~\cite{li2022learning}).

Biases in question answering dataset has long been a problem. In particular, for template-based open-ended question, answers imbalance, where a specific answer category takes up the majority of the question. Such biases can skew models towards certain answer categories, ignoring the actual context. , e.g., in VQA, no visual context. ~\cite{zhang2016yin} verified and mitigated such issue by constructing a balanced synthetic VQA dataset. Such bias issue is further examined in VQA-CP~\cite{agrawal2018dont} by crafting a dataset where answer distribution varies significantly across train and test splits such that models that overfit to specific answer category during training will be `trapped' on the test set. In addition, different models that aim to handle the bias issue were proposed by reducing the language bias and grounding on visual modality~\cite{cadene2020rubi, kv2020reducing}. As audio-visual question answering emerged as a new task, few attentions have been focused on studying the bias issue. In particular for open-ended AVQA dataset, Pano-AVQA~\cite{yun2021panoavqa} only briefly mentioned that bias are alleviated by audio replacement or providing counter examples with different answer to same question. MUSIC-AVQA~\cite{li2022learning}, as the largest and most popular open-ended AVQA benchmark, completely ignores the bias issue presented in each question category. As such, our endeavor to solve bias issue in this particular dataset becomes valuable.

\section{Dataset balancing}
\begin{figure*}[t]
    \centering
    \includegraphics[width=1.0\linewidth]{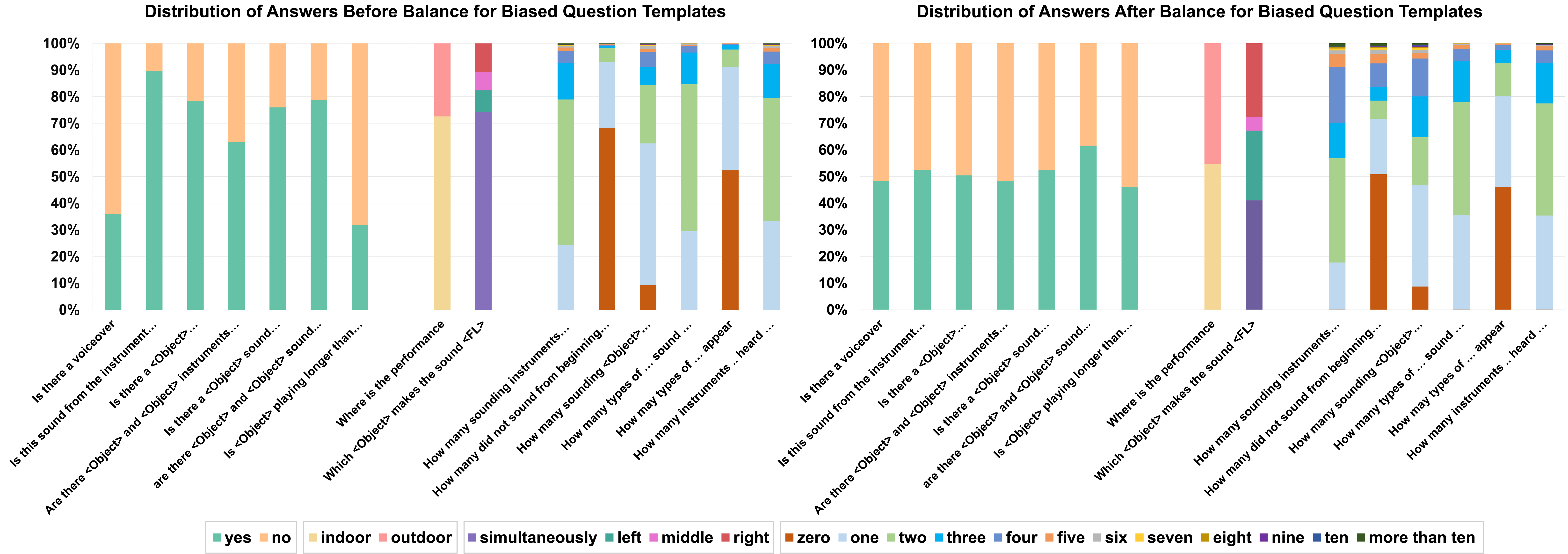}
    \caption{An overview of answer distribution in bias question templates before and after balance.}
    \label{fig:bias_balance_distrib}
\end{figure*}

We describe our approach to balance the dataset on the public MUSIC-AVQA dataset. This process entails two phases: pinpointing biased questions and then achieving balance.
First, we scrutinize the distribution of answers for each question template. Specifically, we review questions templates under all the 9 questions types.
Our criteria for identifying bias is when a single answer represents over 60\% of responses for binary questions or exceeds 50\% for multi-class questions (with more than two possible answers). Out of the 9 question types evaluated, 7 exhibited a skewed answer distribution for at least one of their templates. These include: audio-visual existential, audio-visual counting , audio-visual temporal, visual location, visual counting, audio counting and audio comparative questions. Within these types, multiple templates could exhibit biases, amounting to 15 out of 33 templates in total. A detailed comparison of the answer distribution before and after our balancing for each biased template is illustrated in Fig.~\ref{fig:bias_balance_distrib}. We present the balancing process below.

\textbf{Audio-visual Existential Questions:} Binary questions dominate this question type, where the answers are either ``yes" or ``no". In this case, we simply pick the most frequent answer and collect complementary pairs for it. We take two questions as an example:
\textbf{Is this sound from the instrument in the video?}
90\% of data samples in this question template are answered ``yes". To create QA pairs whose answers are ``no", we replace the audio track from the video with audio of another instrument type. To make the QA pairs non-trivial, we cluster the set of instruments into ``string instrument", ``woodwind instrument", ``brass instrument" and ``percussion instrument". When replacing the audio track, by 50\% chance the audio track is replaced with a different instrument of the same cluster, while another 50\% chance the audio track is replaced with instrument music belonging to other clusters. Using this method, we create 794 videos paired with non-matching audio segments.
\textbf{Is there a voiceover?} Originally, this question is severely imbalanced where 79.6\% of answers are ``no". However, after carefully delving deep into the video data, we found that different individuals have different definitions to the concept of ``voiceover", which led to inconsistencies in the labels. For example, some labelers define ``voiceover" as human voice appearing on top of the instrument sound, while others define it as any general ``off-screen" sound.
To fix such inconsistency, we define it as \textbf{any ``off-screen sounds''}. After manually checking 1,278 video-audio-question pairs from the training set, we corrected 169 mislabeled entries (13\%). Despite this correction, a significant imbalance remained with 68\% of labels still being no". To address this, we add another 456 QA pairs from our collected videos where ``voiceover" presents (answered ``yes"), resulting in a balanced distribution with 51.7\% yes" and 48.3\% no" answers.

\textbf{Audio-Visual Counting Questions:} This type of question asks about counting aspect of instruments in the video. The questions are structured using 4 templates that address the following aspect of counting: (i). the total number of sounding instruments (T1). (ii). the number of instrument types which are sounding (T2), (iii). the number of a specific sounding instrument type (T3), iv). and number of instruments that did not sound from the beginning to the end (T4). The answers are restricted to ``0-10" and ``more than ten".
Upon analyzing the dataset, we observe a significant imbalance in the answers. Specifically, ``0", ``1" and ``2" dominates the answers. For all four templates, the most frequent answer exceeded 50\% of the total, with one template even reaching 60\%.

To balance audio-visual counting questions, we manually collect musical ensemble performance videos. Our goal is to gather videos of musical ensemble performances where at least one answer to the aforementioned question templates exceeded 2. We sourced potential videos from the YouTube8M~\cite{abuelhaija2016youtube8m} dataset, specifically targeting those tagged with terms like "musical ensemble," "string quartet," or specific instrument names. For each instrument type, we selected videos tagged with that type and combined them with videos tagged as "musical ensemble," "string quartet," "quartet ensemble," or a specific instrument name, which helps narrow down potential candidates. From this set, we manually filtered out videos that were of very low quality, had static scenes like album covers, or had ambiguous content.
For each selected video, we annotated:
(i). Total number of instruments,
(ii). Number of distinct instrument types,
(iii). Count of each instrument type,
(iv). Number of sounding instruments,
(v). Number of distinct sounding instrument types,
(vi). Count of the most frequently appearing instrument that also produces sound,
(vii). Number of instruments that did sound from the beginning to the end of the video.
In total, we collected 591 videos for audio-visual counting questions. Using the annotations from these videos, we generated additional QA pairs for each template: 572 (+39\%) for T1, 502 (+25.4\%) for T2, 815 (+40.1\%, 350 from originally unlabeled videos for the question template in MUSIC-AVQA dataset, 465 from our collected videos) for T3, and 413 (+30.3\%) for T4. These new pairs have answers from the less frequent answer categories. After adding these pairs, imbalance issue from all 4 question templates is sufficiently mitigated, with the most frequent answer percentages decreasing by 16\%, 17\%, 15\%, 13\%  for the 4 templates, respectively.

\textbf{Audio-Visual Temporal Questions:} This category of questions probes the order in which instruments play during a performance. The candidate answers range from 22 instrument categories, as well as positional indicators like ``left", ``middle", ``right" that specify an instrument's location. Additionally, the term``Simultaneously" denotes that instruments play at the same time. Among 3 question templates in this question category, the question ``Which  \texttt{<Object>} makes the sound first/last?" shows a strong imbalance: 74\% of answers are ``simultaneously".
To address this imbalance in a multi-class setting, we labeled QA pairs with the answers ``left," ``right," or ``middle" to diminish the dominance of the ``simultaneously" category. For example, consider a video where three violinists are performing. If the violinist on the left initiates the performance, followed by the middle and right violinists, we can formulate a QA pair as: "Q: Which violin starts playing first? A: left." In addition, we augment the video by horizontally flipping it. This transforms the QA pair to: "Q: Which violin starts playing first? A: right." Following the above procedure, we collected 203 additional targeted videos from YouTube for creating QA pairs. After augmentation, we end up creating 713 (+81.1\%) additional QA pairs with answers rather than ``simultaneously", reducing the most frequent answer percentage by 33\% from 74\% to 41\%.

\textbf{Visual Counting Questions:} Unlike audio-visual questions, these questions rely solely on visual information. The first two templates, ``Is there \texttt{<Object>} in the entire video'' and ``Are there \texttt{<Object>} and \texttt{<Object>} instruments in the video'' determine the presence of specific instruments. For these templates, the majority of answers are "yes", constituting 78.4\% and 62.7\% respectively. To counter this bias, we generated 794 and 423 QA pairs with the answer "no" for each template. These pairs were created using labels from our collected videos.

The third template focuses on counting the types of instruments present in the video. There's a notable imbalance here, with the answers ``1" and ``2" making up 91\% of all responses. To address this, we used labels from the 591 videos from our audio-visual counting collection. We specifically chose 204 videos where the answer to this question exceeded 2. This selection helped reduce the dominance of the top two answer categories, bringing their combined percentage down from 91\% to 80\%.

\textbf{Visual Location Questions:} This category of questions pertains to the location of the performance and the specific positioning of a performer. It seeks answers to whether the performance is indoor or outdoor and inquires about the relative position of instruments in the video. Out of the four question templates in this category, the template ``Where is the performance?" with answers either ``indoor'' or ``outdoor'', exhibits an outstanding imbalance. To address this, we collect 456 QA pairs whose answer is ``outdoor'', from 456 videos that are not labeled for this question template from original MUSIC-AVQA dataset, resulting in reduction of  QA pairs with the dominant category (``indoor'') by 17.9\% from 72.6\% to 54.7\%.

\textbf{Audio Counting Questions:} The first two question templates, "Is there a \texttt{<Object>} sound?" and "Are there \texttt{<Object1>} and \texttt{<Object2>} sounds?", focus on the audio aspect, determining the presence of specific sounds. For these templates, the majority of answers are "yes", accounting for 76.0\% and 78.8\% respectively. To address this imbalance, we created 794 and 423 QA pairs with the answer ``no" for each template, using labels from videos where the sounding instruments were identified. This reduced the dominance of the "yes" answer to 52.5\% and 61.5\% for the two templates, respectively.
The third template queries the total number of distinct instrument sounds heard throughout the video. Notably, the answers "1" and "2" represent over 89\% of all responses, highlighting a significant imbalance. To address the issue, we labeled 572 QA pairs from our collection of ensemble performance videos, specifically choosing videos where the answer to this question was neither "1" nor "2". After the balancing, We reduced the combined dominance of the "1" and "2" answers by 28\% to 61\%.

\textbf{Audio Comparative Questions:} This type of question compares different instruments sounding in the video in terms of loudness, duration and rhythm aspect. Among 3 question templates, a question template asking which instrument playing longer has a strong imbalance, where the answer ``no'' (indicating that neither instrument plays significantly longer) represents 68\% of the data. To address this imbalance, we curated 182 QA pairs from previously unlabeled videos in the original dataset, ensuring that the first \texttt{<Object>} plays longer than the second \texttt{<Object>}. This reduces the dominance of ``no'' answer to 53.9\%.

\section{Approach}
\label{sec:model}
\begin{figure*}[t]
    \centering
  \includegraphics[width = \linewidth]{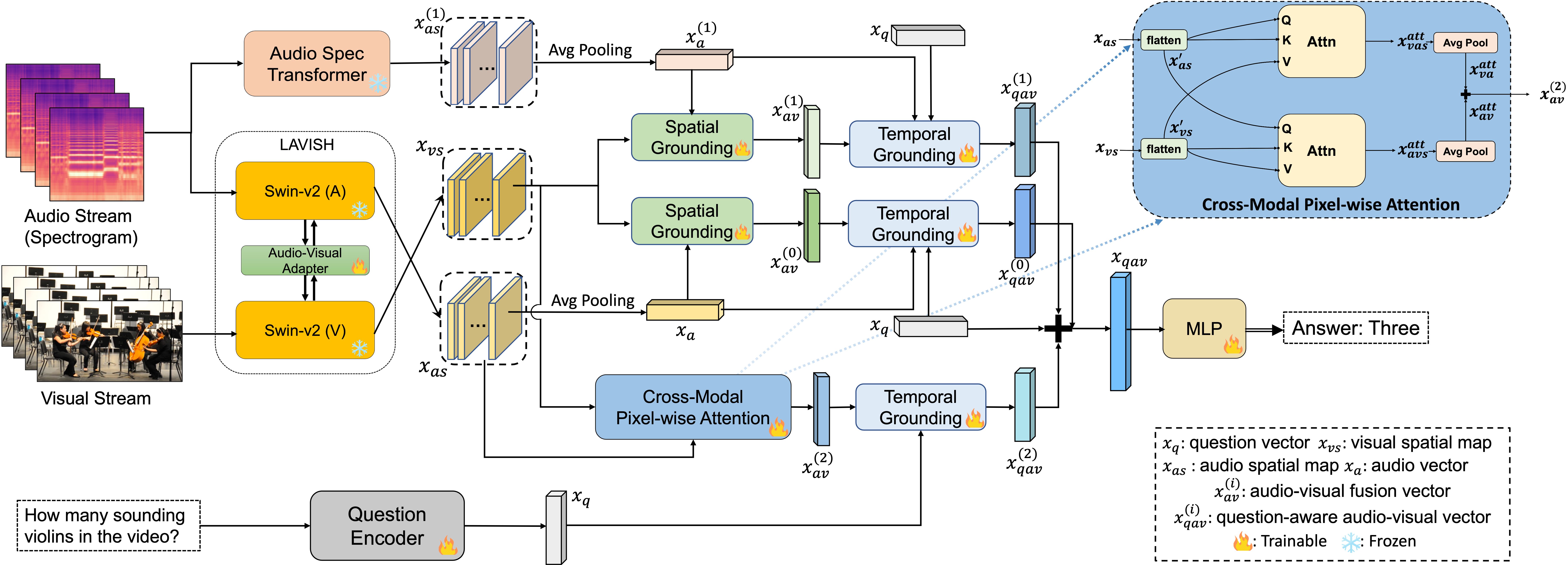}
  \caption{An overview of the new baseline model: (i). A pre-trained Audio-Spectrogram-Transformer (AST) branch is incorporated as an additional audio feature branch. (ii). A cross-modal pixel-wise attention module between audio and visual feature maps is designed to better capture the audio-visual correspondence at a granular level.}
  \label{fig:avst_ensemble}
\end{figure*}

In addition to data balancing, we introduce a new model designed to set a robust baseline for MUSIC-AVQA v2.0. This model integrates existing audio-visual learning and QA components: (i). LAVISH~\cite{lin2023vision}, a 2-tower pretrained Swin-Transformer V2~\cite{liu2022swin} with ``Audio-Visual Adapter" for audio-visual fusion. (ii). A spatial \& temporal grounding module from AVST~\cite{li2022learning}. Our extension to the existing model include:
(i). A new audio-visual fusion branch, the ``AST branch". This extracts audio feature from a pretrained Audio-Spectrogram-Transformer (AST) and merges them with the visual feature branch from Swin-v2 backbone, leveraging audio-visual spatial and temporal grounding~\cite{li2022learning}.
(ii). A cross-modal pixel-wise attention module, further refining audio-visual fusion at a granular level. A detailed overview of the model is illustrated in Fig.~\ref{fig:avst_ensemble}.

\textbf{AST branch:} Incorporating the pretrained audio feature as an auxiliary branch enables our model to capture richer semantic audio information compared to the vision pretrained transformer applied to the audio spectrogram in LAVISH. Specifically, we extract the final hidden output of the AST model, which is a spatial feature map $x^{(1)}_{as}$. We then apply the same operations as those used for the LAVISH~\cite{lin2023vision} branch outputs, namely spatial and temporal grounding as introduced by AVST~\cite{li2022learning} (please refer to it for details).  The feature map $x^{(1)}_{as}$ undergoes average pooling to produce a vector $x^{(1)}_{a1}$, which subsequently computes spatial attention in relation to the visual map, $x_{vs}$
, output from LAVISH's vision branch. The resulting spatial grounding module output is a vector in each frame, $x^{(1)}_{av}$, capturing visual features attended by audio. Following this, temporal grounding is performed. $x^{(1)}_{av}$ is concatenated with audio features, which are then attended by the question vector produced by a LSTM encoder, $x_q$, along the temporal axis. The output of temporal grounding module is a question-aware audio-visual fusion feature $x^{(1)}_{qav}$. This feature is concatenated with outputs from other branches and subsquently passed to a MLP for classification.

\textbf{Cross-modal pixel-wise attention module:} Existing spatial grounding module~\cite{li2022learning} uses a mean-pooled audio vector to compute attention with visual spatial maps. This attention approach: i) Losing the spatial details of spectrogram features. ii) is uni-directional, where only the audio vector serves as a query to the visual maps without any reciprocal interaction. To address these limitations, we propose a refined pixel-wise cross attention between the visual and audio maps, aiming to capture the correspondence between these two modalities more effectively. Specifically, given $x_{vs} \in \mathbb{R}^{H \times W \times C}$ representing frame-level visual map output from LAVISH, and $x_{as} \in \mathbb{R}^{H \times W \times C}$ representing spectrogram frame-level feature output from LAVISH, we compute two pixel-wise audio-visual attentions between two maps. We first flatten the spatial dimension of both feature maps to be $(HW) \times C$, resulting in $x'_{vs}$ and $x'_{as}$. Then we compute mutual cross-attention between these two flattened maps, where each map attends to the other :
 \begin{align}
 & x^{att}_{vas} = x'_{as} + \text{Softmax}(\frac{x'_{vs} x'^{T}_{as}}{HW})x'_{as},  \\
 & x^{att}_{avs} = x'_{vs} + \text{Softmax}(\frac{x'_{as} x'^{T}_{vs}}{HW})x'_{vs}
\end{align}
The obtained $x_{vas}$ and $x_{avs}$ represent the pixel-wise fusion maps for audio and vision branch. We then normalize both maps and average pool their spatial dimensions to produce two vectors, $x_{av}$ and $x_{va}$. These are concatenated to yield the final module output, $x^{(2)}_{av}$. $x^{(2)}_{av}$ is same dimension as $x^{(0)}_{av}$, $x^{(1)}_{av}$ - outputs of the spatial grounding modules from the other 2 branches. As a result, subsequent operations, such as temporal grounding, remain consistent with the other two branches.

\section{Experiments}
\subsection{Models Evaluation on Bias and Balanced Dataset}
To further verify the data bias existing in the MUSIC-AVQA dataset and its negative impact to the model performance, we evaluate AVST and LAVISH models on two newly created datasets (biased and balanced datasets) here.
We first create a balanced test set by sampling 20\% from MUSIC-AVQA v2.0, with balanced answer distribution in every question categories using stratified sampling. Then within this balanced test set, we sample a bias test set by keeping the same QA distribution as MUSIC-AVQA dataset~\cite{li2022learning} ( with our corrected QA pairs in ``voiceover" category). Both the biased and balanced set test are used to evaluate all models.

\begin{figure*}[t]
    \centering
  \includegraphics[width = 0.75\linewidth]{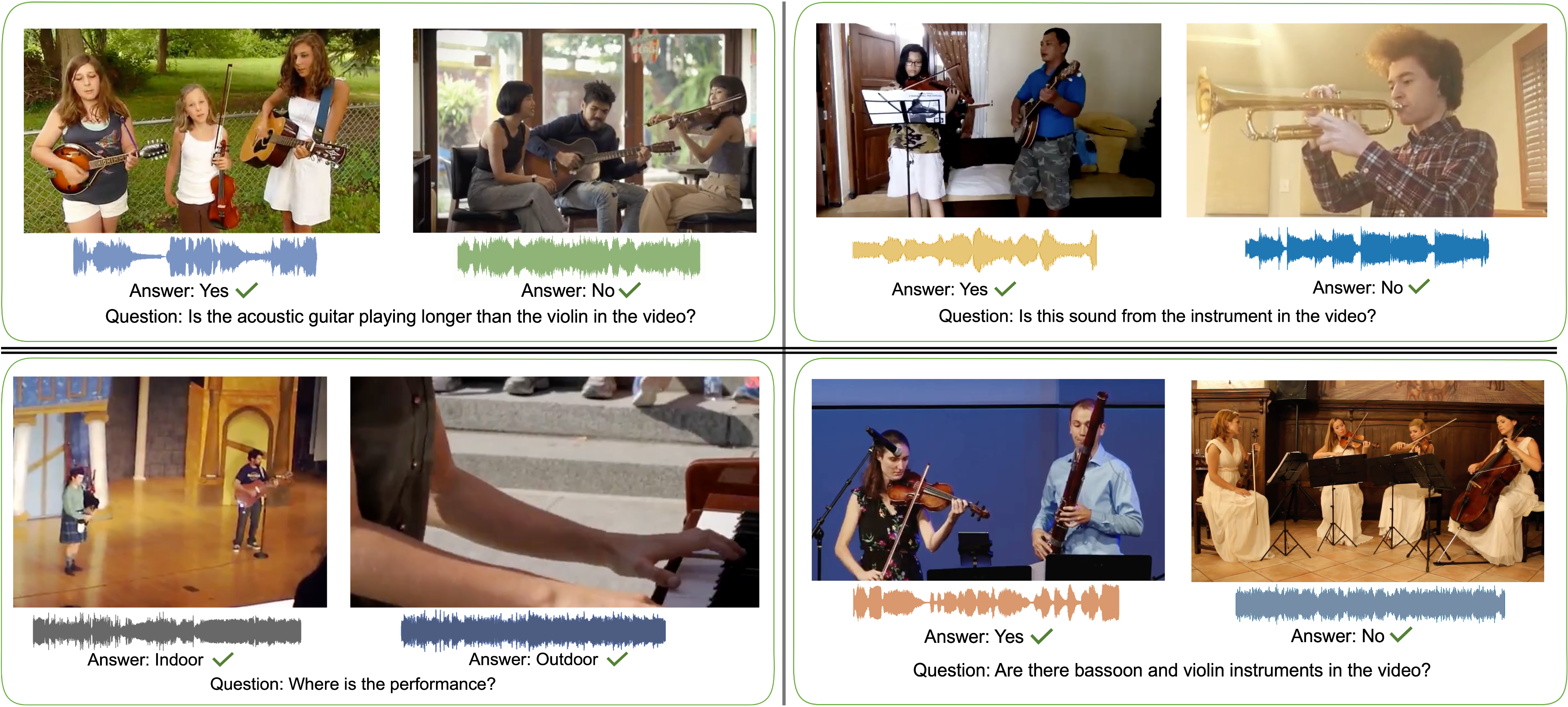}
  \caption{Qualitative Results of `LAST-Att' model on contrasting binary QA samples from different question categories. The model can correctly predict the pairs of QA with same question but opposite answers.}
  \label{fig:qual}
\vspace{-3mm}
\end{figure*}

After reserving the balanced test set from MUSIC-AVQA v2.0, the residual data remains balanced. To ensure a fair comparison, it is imperative that the training sets for both the bias and balanced datasets are of equal size. To achieve this, we first extract a bias subset from the leftover balanced data. To maximize the training samples in this bias set, any QA pair with an answer belonging to the most frequent answer category within its question template is incorporated into the biased subset.

Once the balanced test set is held out from MUSIC-AVQA v2.0, the remaining part is still a balanced dataset. To guarantee the fairness of comparison, we need to keep the size of training set same for the bias and the balanced set. To achieve this, we first sample the bias subset from the remaining balanced set. In order to maximize the number of training samples for the bias set, any QA pair whose answer is the most frequent answer category in its question template is included into the bias subset. For QA pairs in remaining answer categories in the question template, the numbers are determined by: $N_{most} \times \frac{N^{'}_{other}}{N^{'}_{most}}$, where $N_{most}$ is the number of QA pairs whose answer is the most frequent answer category of the question template in our remaining balanced set,  $N^{'}_{most}$ is the number of QA pairs whose answer is the most frequent answer category of the question template in MUSIC-AVQA training dataset, and $N^{'}_{other}$ is the number of QA pairs whose answer is from another less frequent category of the question template in MUSIC-AVQA training dataset. Once we create the bias subset, we sample another balanced subset by keeping the same number of samples as the bias subset. In the last step, we reserve $1/8$ for validation, with the remaining $7/8$ forming our biased and balanced training sets.

\begin{table*}[t]
\vspace{-2mm}
    \centering
    \caption{Evaluate Existing Models on Balanced Test set. Highlights are results where both models trained on balanced set consistently outperform trained on bias set. (Ext: Existential. Cnt: Counting. Temp: Temporal. Comp: Comparative.)}
    \resizebox{0.7\linewidth}{!}{%
    \begin{tabular}{l|l|c|ccccc|cc|cc}
        \toprule
        \multirow{2}{*}{Model} & \multirow{2}{*}{Train set} & \multirow{2}{*}{Total} & \multicolumn{5}{c|}{Audio-Visual} & \multicolumn{2}{c|}{Visual} & \multicolumn{2}{c}{Audio} \\
        & & & Ext & Temp & Cnt & Loc& Comp & Cnt & Loc & Cnt & Comp \\
        \hline
        \multirow{2}{*}{AVST~\cite{li2022learning}} & bias & 69.40 & 70.16 & 61.94 & 62.99 & 63.26 & 63.94 & 75.02 & 78.85 & 77.86 & 63.61 \\
         & balanced & \textbf{71.15} & \textbf{71.38} & 59.98 & \textbf{68.85} & 63.48 & 65.40 & \textbf{77.61} & 77.80 & \textbf{82.13} & 62.05 \\ \hline
        \multirow{2}{*}{LAVISH~\cite{lin2023vision}} & bias & 70.39 & 68.11 & 60.08 & 66.72 & 65.11 & 63.12 & 78.14 & 81.46 & 78.80 & 60.96 \\
         & balanced & \textbf{73.35} & \textbf{72.04} & 63.19 & \textbf{71.95} & 67.07 & 63.40 & \textbf{80.40} & 82.66 & \textbf{83.57} & 63.14 \\
        \bottomrule
    \end{tabular}%
    }
    \label{tab:balance_test}
\end{table*}
\begin{table*}[t]
    \centering
    \caption{Evaluate Existing Models on Bias Test set (Ext: Existential. Cnt: Counting. Temp: Temporal. Comp: Comparative.)}
    \resizebox{0.7\linewidth}{!}{%
    \begin{tabular}{l|l|c|ccccc|cc|cc}
        \toprule
        \multirow{2}{*}{Model} & \multirow{2}{*}{Train set} &  \multirow{2}{*}{Total} & \multicolumn{5}{c|}{Audio-Visual} & \multicolumn{2}{c|}{Visual} & \multicolumn{2}{c}{Audio} \\
        & & & Ext & Temp & Cnt & Loc & Comp & Cnt & Loc & Cnt & Comp \\
        \hline
        \multirow{2}{*}{AVST~\cite{li2022learning}} & bias & \textbf{73.07} & \textbf{85.68} & 66.02 & 69.97 & 63.26 & 63.94 & 77.50 & 77.89 & 83.56 & 65.57 \\
         & balanced & 72.01 & 75.36 & 64.68 & 70.82 & 63.48 & 65.40 & 77.91 & 76.75 & 84.44 & 62.11 \\ \hline
        \multirow{2}{*}{LAVISH~\cite{lin2023vision}} & bias & \textbf{74.59} & \textbf{84.79} & 67.84 & 73.53 & 65.11 & 63.12 & 80.77 & 81.14 & 84.54 & 63.59 \\
         & balanced & 74.00 & 75.36 & 68.33 & 73.37 & 67.07 & 63.40 & 80.36 & 81.79  & 84.83 & 63.92 \\
        \bottomrule
    \end{tabular}%
    }
    \label{tab:bias_test}
\end{table*}
\begin{table}
    \centering
    \caption{Evaluation Results on Balanced Test Set: Our new baselines v.s existing models. (Ext: Existential. Cnt: Counting. Temp: Temporal. Comp: Comparative.)}
    \resizebox{\linewidth}{!}{%
    \begin{tabular}{l|c|ccccc|cc|cc}
        \toprule
        \multirow{2}{*}{Model} &  \multirow{2}{*}{Total} & \multicolumn{5}{c|}{Audio-Visual} & \multicolumn{2}{c|}{Visual} & \multicolumn{2}{c}{Audio} \\
        & & Ext & Temp & Cnt & Loc & Comp & Cnt & Loc & Cnt & Comp \\
        \hline
        AVST~\cite{li2022learning} & 71.02 & 72.44 & 59.36 & 68.22 & 65.54 & 63.31 & 77.48 & 77.88 & 82.34 & 60.81 \\
        LAVISH~\cite{lin2023vision} & 73.18 & 73.83 & 60.81 & 73.28 & 65.00 & 63.49 & 81.99 & 80.57 & 84.37 & 58.48 \\
        LAST & 74.85 & 74.08 & 59.15 & 75.17 & \textbf{69.02} & \textbf{66.12} & 83.19 & 83.41 & 85.75 & 61.59 \\
        LAST-Att & \textbf{75.44} & \textbf{76.21} & \textbf{60.60} & \textbf{75.23} & 68.91 & 65.60 & \textbf{84.12} & \textbf{84.01} & \textbf{86.03} & \textbf{62.52} \\
        \bottomrule
    \end{tabular}%
    }
    \vspace{-0.3cm}
    \label{tab:baseline_balance}
\end{table}
\begin{table}
    \centering
    \caption{Evaluation Results on Bias Test Set: Our new baselines v.s existing models. (Ext: Existential. Cnt: Counting. Temp: Temporal. Comp: Comparative.)}
    \resizebox{\linewidth}{!}{%
    \begin{tabular}{l|c|ccccc|cc|cc}
        \toprule
        \multirow{2}{*}{Model} &  \multirow{2}{*}{Total} & \multicolumn{5}{c|}{Audio-Visual} & \multicolumn{2}{c|}{Visual} & \multicolumn{2}{c}{Audio} \\
        & & Ext & Temp & Cnt & Loc & Comp & Cnt & Loc & Cnt & Comp \\
        \hline
        AVST & 71.92 & 75.36 & 64.81 & 70.51 & 65.54 & 63.31 & 77.74 & 76.67 & 84.74 & 61.78 \\
        LAVISH~\cite{lin2023vision} & 73.51 & 74.92 & 66.5 & 75.08 & 65 & 63.49 & 82.08 & 79.59 & 85.32 & 59.14 \\
        LAST & 75.24 & 75.58 & 65.78 & 76.16 & \textbf{69.02} & \textbf{66.12} & 83.63 & 82.36 & \textbf{86.20} & 61.78 \\
        LAST-Att & \textbf{75.45} & \textbf{76.47} & \textbf{66.75} & \textbf{76.20} & 68.91 & 65.60 & \textbf{83.86} & \textbf{83.09} & 85.71 & \textbf{63.10} \\
        \bottomrule
    \end{tabular}%
    }
    \vspace{-0.3cm}
    \label{tab:baseline_bias}
\end{table}
\begin{table}[h]
    \centering
    \caption{Model Performance on Contrasting Binary audio-visual QA pairs}
    \resizebox{0.45\linewidth}{!}{%
    \begin{tabular}{l|c}
        \toprule
        Model  & Total Accuracy \\
        \hline
        AVST~\cite{li2022learning} & 52.4 \\ 
        LAVISH~\cite{lin2023vision}  & 54.47 \\
        LAST & \textbf{58.00} \\
        LAST-Att & \textbf{58.86} \\
        \bottomrule
    \end{tabular}%
    }
    \label{tab:contrast_qa}
    \vspace{-0.3cm}
\end{table}

By following the outlined procedure, we create 2 datasets, a bias and a balanced set. Each dataset contains 31,513 training samples, mirroring the size of the original MUSIC-AVQA training set (approximately 31k). Additionally, both datasets share 4,502 validation samples, 10,819 balanced test samples, and 9,119 biased test samples. We then assess the performance of 2 existing open-source models, the baseline AVST model~\cite{li2022learning} and the state-of-the-art LAVISH~\cite{lin2023vision} model. Both models are evaluated on the biased and balanced test sets respectively, adhering strictly to their original training guidelines without any modifications. To validate the integrity of our code and data, we trained, validated, and tested both models using the original MUSIC-AVQA dataset. The AVST model achieved a total accuracy of 71.25\% on the test set (compared to the reported 71.51\%), while the LAVISH model achieved 77.17\% (close to the reported 77.20\%). These results align closely with the numbers reported in their publications.

We proceeded with training both models on the bias and balanced training sets, resulting in 4 distinct models. For models trained on the bias set, we use the total accuracy on the bias validation set to select the best checkpoint across epochs. Similarly, for models trained on the balanced set, we use the total accuracy on the balanced validation set for checkpoint selection. After training, we evaluated all four models on both the bias and balanced test splits. The evaluation results are presented in Table ~\ref{tab:balance_test} for the balanced test set and in Table ~\ref{tab:bias_test} for the bias test set. As shown in Table ~\ref{tab:balance_test}, both models trained using the balanced set achieve higher total accuracy on the balanced test set, with gains of \textbf{+1.75\%} for AVST and \textbf{+2.96\%} for LAVISH. Specifically, in question types where severe answer imbalance exists, such as audio-visual counting and existential questions, models trained on our balanced data consistently achieve higher accuracy on the balanced test set. For instance, the LAVISH model showed improvements of \textbf{+3.93\%} and \textbf{+5.23\%} respectively. Conversely, as shown in Table ~\ref{tab:bias_test}, all models trained on the bias set surpassed those trained on the balanced set when evaluated on the biased test set. These results suggest the data bias in original MUSIC-AVQA. Models trained from the data overfit to the biases, thereby undermining their ability to generalize well.

\subsection{New Baseline Evaluation}
To establish new baselines, we evaluate our proposed models using MUSIC-AVQA v2.0, our entire balanced dataset, comprising 36.7k QA pairs for training, 5,250 for validation, and 10,819 for testing. We compare two model variants, `LAST' and `LAST-Att', against existing models. The LAST variant integrates the existing LAVISH model with the AST branch, as detailed in Section ~\ref{sec:model}. LAST-Att represents our full model, incorporating both the AST branch and cross-modal pixel-wise attention. Further implementation specifics for both models can be found in the Supplementary Material. As shown in Table ~\ref{tab:baseline_balance} and Table ~\ref{tab:baseline_bias}, both of our proposed baselines surpass the performance of LAVISH and AVST on both test sets. Notably, on the balanced test set, LAST-Att achieves a performance boost of \textbf{+2.26\%} over LAVISH and \textbf{+4.42\%} over AVST.

\subsection{Contrasting Binary QA Evaluation}
\vspace{-2mm}
To assess the models' true capability in understanding audio-visual contexts and to ensure they do not merely rely on language priors, we design a contrasting binary QA evaluation. We curated a subset from the balanced test split of MUSIC-AVQA v2.0, consisting of paired binary QA samples. Each pair contains two questions with identical phrasing but pertains to different videos that have opposite answers. For instance, the question `Are there ukulele and violin sounds in the video?' is posed for both video A and B. While the answer for video A is ``yes", it's ``no" for video B. From the balanced test set, we identified 1643 such contrasting binary QA pairs. For evaluation, a model must correctly answer both questions in a pair to be deemed accurate for that pair. This ensures that the model truly comprehends the audio and visual context of each video, given that the question phrasing is the same. As shown in Table ~\ref{tab:contrast_qa}, in this challenging scenario, our LAST-Att model strongly outperforms the LAVISH and AVST models trained on the balanced set by margins of +4.39\% and +6.46\%, respectively. Some of the contrasting QA samples predicted by our LAST-Att' model are illustrated in Fig.~\ref{fig:qual}. Overall, our model can correctly predict QA pairs by better reasoning the audio-visual context than relying on language priors.

\section{Conclusion and Discussion}
In this work, we identify the data bias that exists in the public MUSIC-AVQA dataset, which hinders the progress of multimodal research. We systematically address the data imbalance issue, and propose a new and larger balanced dataset, MUSIC-AVQA v2.0. Extensive results show that the strong bias from MUSIC-AVQA negatively affect the model ability, favoring more to dominant answers from training set. However, existing models trained on our balanced dataset do not overfit to the bias distribution and generalize well, proving that MUSIC-AVQA v2.0 is a more reliable benchmark for AVQA task. Furthermore, we propose a new model on MUSIC-AVQA v2.0. With added AST branch and an additional cross-modal pixel-wise attention, it consistently outperforms previous methods on almost every question categories, serving as a strong baseline for AVQA task.

{\small
\bibliographystyle{ieee_fullname}
\bibliography{egbib}
}

\appendix
\section{Implementation Details}
\subsection{Video Pre-processing}
We adhere to the LAVISH open-source code guidelines to preprocess the audio and visual frames of videos in MUSIC-AVQA and v2.0. The majority of these videos have a duration of 60 seconds. For videos shorter than 60 seconds, we extend them by repeating the last visual frame and the corresponding 1-second audio until reaching 60 seconds, as per the LAVISH guidelines. Visual frames are extracted from videos at a rate of 1fps, yielding 60 frames for each video. For audio, we sample the waveforms at a 16kHz rate. Given the large size of audio and visual frames, models cannot process all frames from a video. Thus, we implement the same down-sampling method as LAVISH, extracting every 10th visual frame from the start of the video, each paired with its corresponding 2-second audio segment. After down-sampling, we are left with 10 visual frames and 10 associated audio waveforms for each video. To accommodate the input size of the Swin-Transformer-V2~\cite{liu2022swin} (the backbone of LAVISH), we resize the image of each frame to $192 \times 192 \times 3$. For every 2-second audio segment, we compute the mel-spectrogram using a 5.2-millisecond frameshift and a kaldi fbank with 192 triangular mel-frequency bins. We then triple the mel-spectrogram along the channel dimension, resulting in a tensor of size $192 \times 192 \times 3$, same dimensions as the image input in each frame.

\subsection{LAVISH~\cite{lin2023vision}}
For LAVISH, we use the official open-source code implementations and detail them below. The backbone of LAVISH consists of a 2-tower Swin-Transformer-V2-Large pretrained on ImageNet: one tower for visual input and the other for mel-spectrogram input. Within each layer of the two towers, two LAVISH adapters (a type of adapter~\cite{houlsby2019parameterefficient} for audio-visual learning) are inserted. One is positioned as the residual of the Multi-Head Attention~\cite{vaswani2023attention} module, and the other as the residual of the MLP module. Both the visual and audio branches undergo cross-attention within these adapters. For details about the LAVISH adapter, please refer to the original paper.

The output from both the visual and audio backbones is a $6 \times 6$ feature map with a channel size of 512. The audio feature map is then mean-pooled across its spatial dimensions to produce a 512-dimensional vector. This vector is forwarded to the spatial grounding module, which attends to the visual feature map for audio-visual fusion. The result of the spatial grounding is a 512-dimensional vector for each frame, with a total of 10 frames. This is then directed to the temporal grounding module, which is a single-layer Multi-Head Attention between the 512-dimensional question vector from a 2-layer LSTM~\cite{HochSchm97} encoder and the spatial grounding outputs on the temporal axis. The final output from the temporal grounding module is a single 512-dimensional vector, which is subsequently concatenated with outputs from other branches. In LAVISH, both the spatial and temporal grounding modules are consistent with the methods described in the AVST work~\cite{li2022learning}. For further details, please refer to it.

\subsection{AST branch}
We apply the pretrained checkpoint of Audio-Spectrogram-Transformer (AST) on AudioSet (``Full AudioSet, 10 tstride, 10 fstride, with Weight Averaging (0.459 mAP)") as the backbone for our ``AST" branch. For the audio spectrogram input, we follow the processing steps outlined in the AST paper. Each 2-second audio waveform segment is converted into a series of 128-dimensional log Mel filterbank (fbank) features, using a 25ms Hamming window at 10ms intervals. This results in a $128 x 204$ mel-spectrogram, which is then used as the input for the AST. Additionally, since the original pretrained AST is designed for a 10-second mel-spectrogram input, whose position embeddings are larger than those for a 2-second input,  we adjust it by symmetrically trimming the leftmost and rightmost portions of the embedding matrix. This ensures the position embeddings are compatible with our 2-second inputs.

Once the audio inputs are processed through AST, we extract the hidden states from the final layer, selecting only the last hidden state to produce a 768-dimensional audio vector. A Linear layer is then added to project this audio vector down to 512 dimensions, aligning with the channel size of the visual feature maps in LAVISH. Subsequently, we employ the same grounding operations as LAVISH, using shared weights in the grounding modules. This involves computing the spatial grounding between the audio vector and the LAVISH visual feature maps. The resulting output is then grounded with the 512-dimensional question vector from the LSTM encoder (as in LAVISH) along the temporal axis. The grounding output from the AST branch matches the LAVISH branch in dimension, producing a 512-dimensional vector. This grounding output from the AST branch is concatenated with outputs from other branches to form an ensemble vector.

\subsection{Cross-modal Pixel-wise Attention}
The module receives two feature maps as inputs: an audio spatial map and a visual spatial map. Both maps have dimensions of $6 \times 6 \times 512$. These maps are flattened to $36 \times 512$, and cross-attention is computed between them along the spatial axis, as detailed in Section 4 of the main paper. The module's output is a 512-dimensional vector, which is then directed to the temporal grounding module (with shared weights) for question-related attention, consistent with the other two branches. This temporal grounding output remains a 512-dimensional vector and is concatenated with outputs from the other branches to form an ensemble vector. Finally, the concatenated outputs from all branches are forwarded to a 2-layer MLP with hidden sizes of 512 and 42 (the vocabulary size of candidate answers) respectively, producing the logits of the answer.

\subsection{Training Details} We implement all models using PyTorch~\cite{paszke2019pytorch}. For LAVISH and AVST, we train the models using cross-entropy loss between the predicted and the ground truth answers, along with an audio-visual matching loss by sampling non-matching visual frames from other videos, as proposed in AVST~\cite{li2022learning}. Following AVST and LAVISH, we assign a weight of 0.5 to the audio-visual matching loss and 1.0 to the cross entropy loss. For our ``LAST" and ``LAST-Att" models, we use cross entropy loss only without audio-visual matching loss. During training, we freeze the parameters of all backbones, including the 2 Swin-Transformers in LAVISH and the AST audio encoder. For LAVISH adapters, we follow the paper to set a small learning rate of 8e-5. And we set the learning rate of 3e-6 for the grounding modules including our cross-modal pixel-wise attention module, and the final prediction layer in AVQA. We use Adam~\cite{kingma2017adam} optimizer to train all models. In terms of hardware configuration, models are trained on 8 NVIDIA-V100 32GB GPUs in data parallel mode. We configure the batch size to 24 for data loading.

\section{Data Collections and Statistics}
\subsection{Data Quality Control}
To ensure the quality of our collected data, we annotate all labels in conjunction with QA pairs by ourselves. Prior to data collection, we meticulously review the QA pairs across 33 templates. This helps us accurately understand the questions and their corresponding videos, minimizing inconsistent annotations stemming from misunderstandings. Upon review, we discover significant inconsistency in a predominant question template within the Audio-Visual Existential category: ``Is there a voiceover". This inconsistency arises from varying interpretations of the term ``voiceover" by previous annotators in MUSIC-AVQA. From the labels, it is evident that some annotators perceived a human voice layered over instrument sounds as a voiceover, while others interpreted it as a generic ``off-screen sound". For consistency, we adopt the latter definition for our annotations. As a result, we adjust 13\% of the annotations from this question template in the training set. For each QA pair among our additionally collected 8,136 samples, we have three individuals verify the annotation and only accept those that received unanimous agreement.

\subsection{Details of Distribution After Balance}
\begin{figure}[t]
    \centering
    \includegraphics[width=0.75\linewidth]{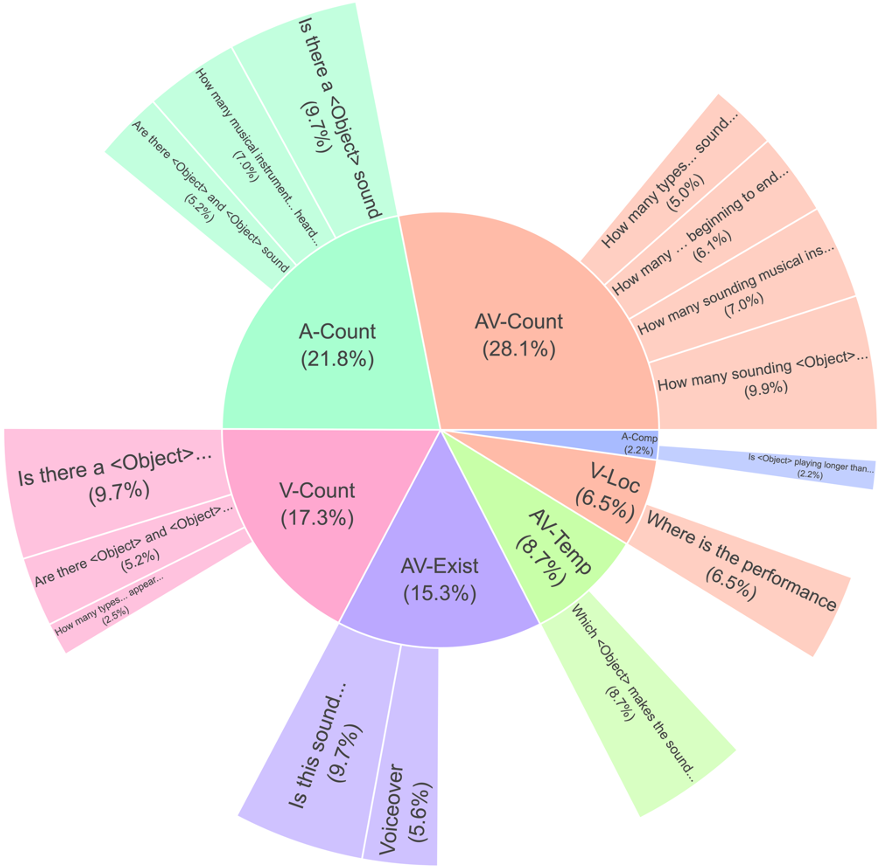} 
    \caption{An overview of the distribution of our collected QA pairs in each question type and question template.}
    \label{fig:distrib_collect_QA}
\end{figure}

\textbf{Additional QA data} We provide detailed statistics for our additional QA pairs. As shown in Fig.~\ref{fig:distrib_collect_QA}, among our 8.1k QA pairs (+17.8\% additional QA pairs to 45.6k in MUSIC-AVQA~\cite{li2022learning} (updated version in their work)), Audio-Visual questions constitute 52.1\% of the total. This includes Audio-Visual Existential at 15.3\%, Audio-Visual counting at 28.1\%, and Audio-Visual Temporal at 8.7\%.  Visual questions account for 23.8\%, with Visual Counting 17.3\% and Visual Location 6.5\%. Audio questions comprise 24\%, with Audio Counting 21.8\% and Audio Comparative 2.2\%.

\begin{figure}[t]
    \centering
    \includegraphics[width=0.85\linewidth]{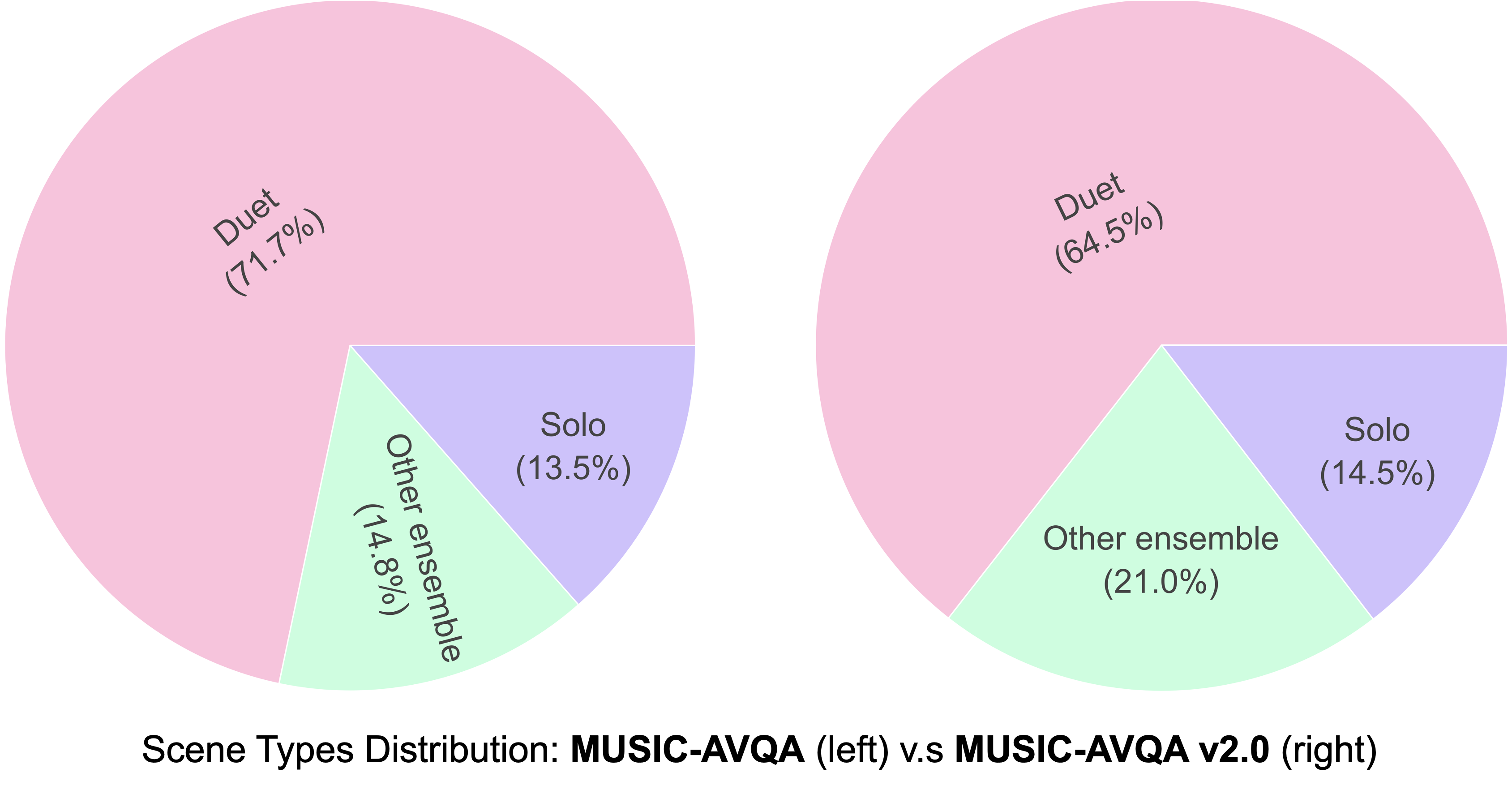} 
    \caption{Distribution of Scene Types}
    \label{fig:scene_type}
\end{figure}

\textbf{Additional Video Data} We collect 1204 additional real videos (+16.2\% more real videos than MUSIC-AVQA (7422)), sourcing from YouTube and YouTube-8M~\cite{abuelhaija2016youtube8m}. Among 1204 videos, 715 videos contain 3 or more instruments,  238 are duets and the remaining 251 are solos. With these additional videos mainly focusing on musical ensembles, we enrich the dataset with more diverse videos and a less skewed distribution of scene types. As illustrated in Fig.~\ref{fig:scene_type}, after collection, the proportion of other ensembles has seen a large increase of 6.2\%, moving from 14.8\% to 21.0\%.

\begin{figure*}[t]
    \centering
    \includegraphics[width=0.8\linewidth]{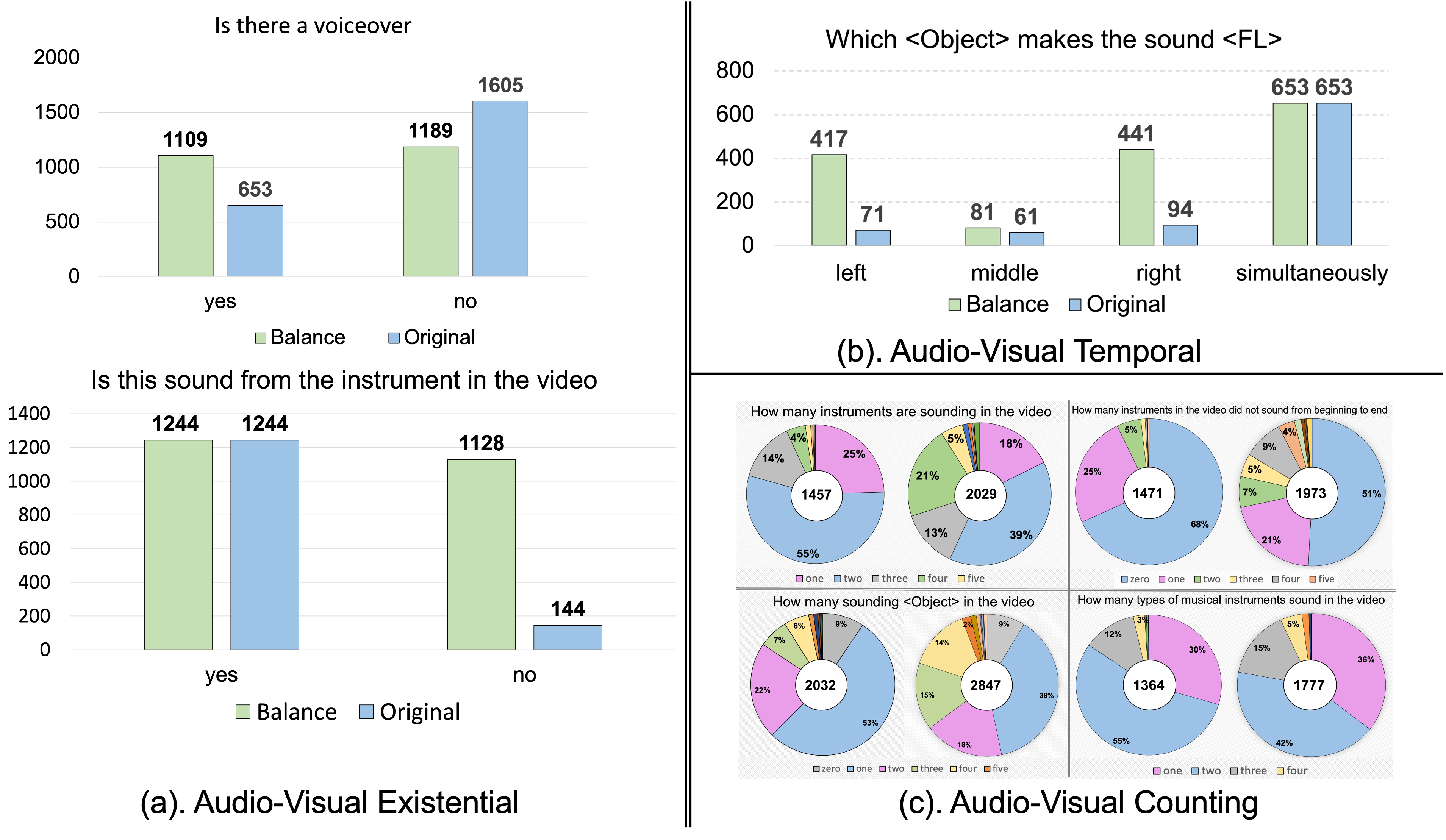} 
    \caption{Distribution of Bias Audio-Visual Questions Before and After Balance}
    \label{fig:audio-visual}
\end{figure*}

\begin{figure}[t]
    \centering
    \includegraphics[width=\linewidth]{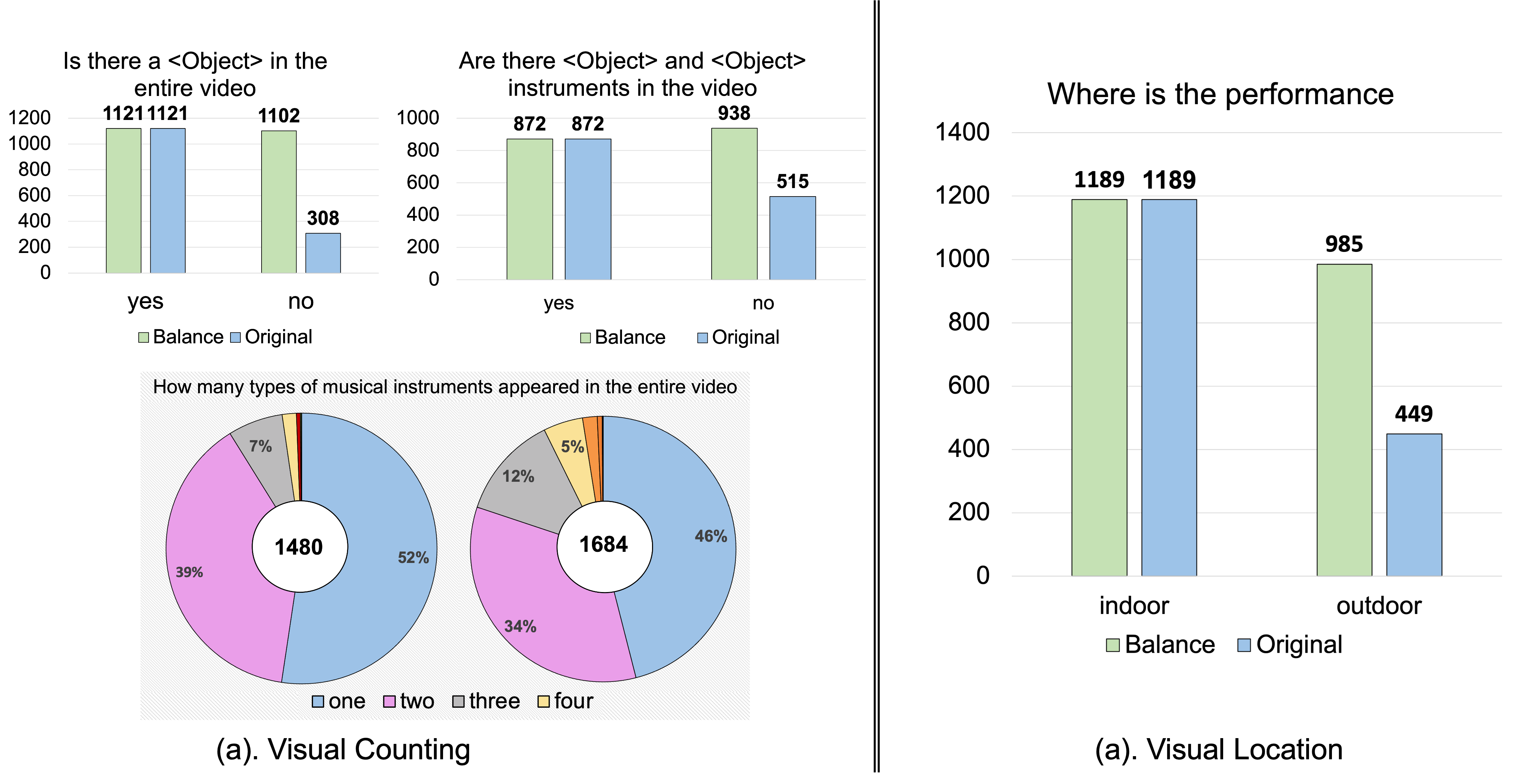} 
    \caption{Distribution of Bias Visual Questions Before and After Balance}
    \label{fig:visual}
\end{figure}

\begin{figure}[t]
    \centering
    \includegraphics[width=\linewidth]{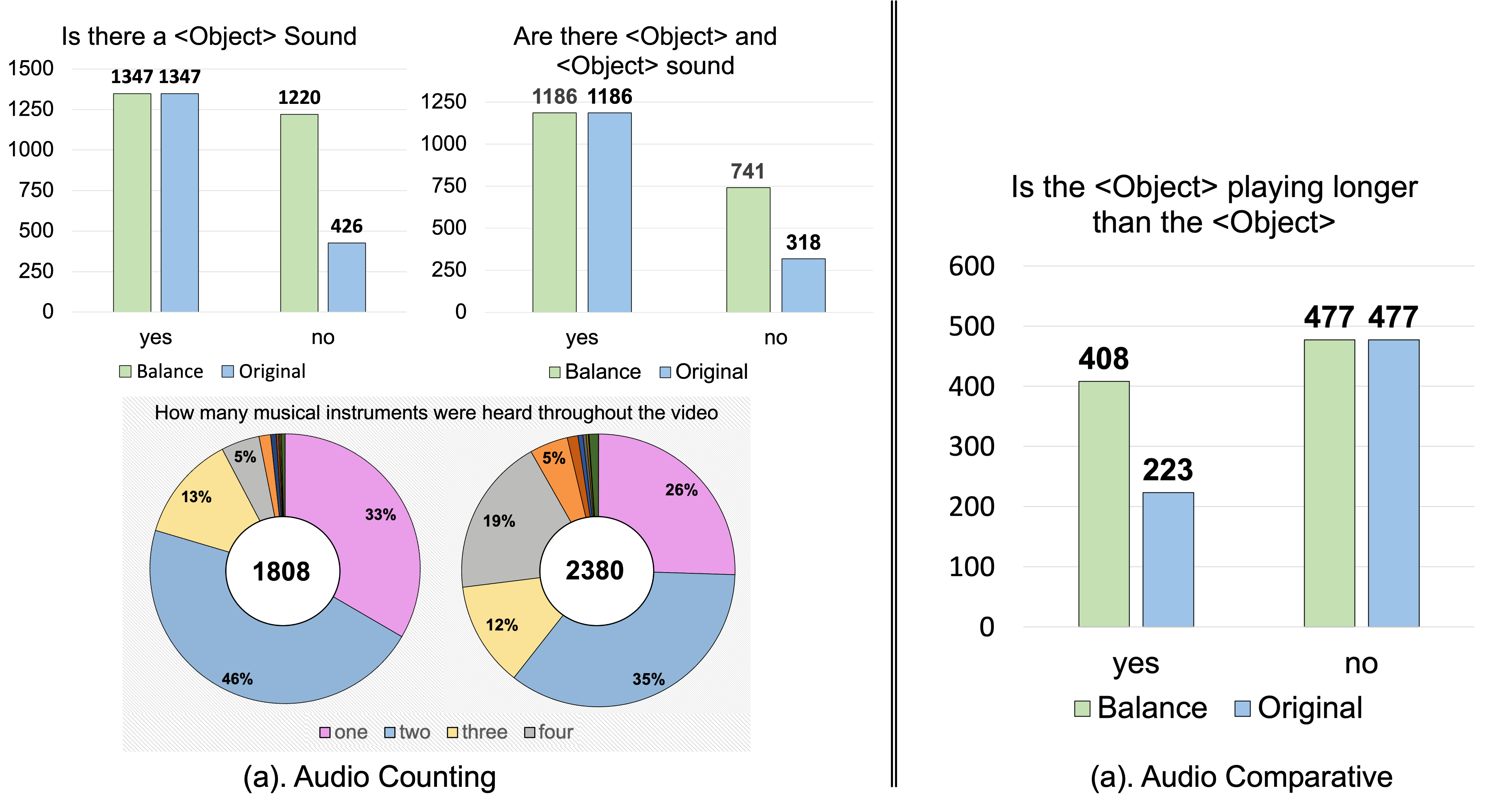} 
    \caption{Distribution of Bias Audio Questions Before and After Balance}
    \label{fig:audio}
\end{figure}

\textbf{Distribution of Each Bias Question Template Before and After Balance} We further show distribution of each bias question template before and after balance, grouped by question type and modality type, Fig.~\ref{fig:audio-visual}, Fig.~\ref{fig:visual} and Fig.~\ref{fig:audio} show Audio-Visual questions, Visual questions and Audio questions respectively. Within each figure, we show specific counts for each answer category within the templates. As evident from the data, our QA collection considerably rectifies the majority of the skewed distribution present in the original dataset. However, we acknowledge that certain counting question templates remain biased towards fewer counts. For instance, questions asking about the number of distinct instrument types inherently lean towards smaller counts, which makes the data collection of larger counts very challenging.

\section{Ablation Studies}
To study the effectiveness of our proposed modules, the ``AST branch" and ``Cross-Modal Pixel-wise Attention", we conduct an ablation study. In this study, we implement two additional models that exclude these proposed modules.The first model, termed ``Swin-AST" model, retains only the vision branch of the LAVISH backbone, excluding the audio branch and the LAVISH adapter. This model can be viewed as an advanced version of the ``AVST" model, given that it substitutes more robust backbones (In AVST, vision branch uses ResNet-18~\cite{he2015deep} and audio branch uses Vgg-ish~\cite{45611} pretrained on AudioSet). The second model, ``LAVISH-Att", preserves just the LAVISH components (the 2-tower backbone, spatial grounding, and temporal grounding) and integrates our cross-modal pixel-level attention module. We train and validate both models on MUSIC-AVQA v2.0, and evaluate on the balanced test split. The ablated model components are detailed in Table~\ref{tab:component_ablation}, and evaluation results are summarized into Table~\ref{tab:ablation_balance}.

\begin{table}
    \vspace{5mm}
    \centering
    \caption{Component Ablation Overview: A breakdown of components used in each method. A (\checkmark) indicates the component is included in the method, while a (-) indicates its absence. ``Swin-A" denotes the LAVISH audio branch, ``AST" denotes the AST audio encoder, ``CM-P-Attn" denotes cross-modal pixel-wise attention module.}
    \resizebox{0.6\linewidth}{!}{%
    \begin{tabular}{l|c|c|c}
        \toprule
        Model & Swin-A & AST & CM-P-Attn \\
        \hline
        LAVISH~\cite{lin2023vision} & \checkmark & - & - \\
        Swin-AST & - & \checkmark & - \\
        LAST & \checkmark & \checkmark & - \\
        LAVISH-Att &\checkmark & - & \checkmark \\
        LAST-Att & \checkmark & \checkmark & \checkmark \\
        \bottomrule
    \end{tabular}%
    }
    \vspace{-0.3cm}
    \label{tab:component_ablation}
\end{table}

\begin{table}
    \centering
    \caption{Ablation Study: Evaluation Results on Balanced Test Set: ``Swin-AST" and ``LAVISH-Att" v.s. Our 2 baselines and existing methods. (Ext: Existential. Cnt: Counting. Temp: Temporal. Comp: Comparative.)}
    \resizebox{1.05\linewidth}{!}{%
    \begin{tabular}{l|c|ccccc|cc|cc}
        \toprule
        \multirow{2}{*}{Model} &  \multirow{2}{*}{Total} & \multicolumn{5}{c|}{Audio-Visual} & \multicolumn{2}{c|}{Visual} & \multicolumn{2}{c}{Audio} \\
        & & Ext & Temp & Cnt & Loc & Comp & Cnt & Loc & Cnt & Comp \\
        \hline
        LAVISH~\cite{lin2023vision} & 73.18 & 73.83 & 60.81 & 73.28 & 65.00 & 63.49 & 81.99 & 80.57 & 84.37 & 58.48 \\
        Swin-AST & 74.63 & 75.88 & 61.84 & 74.31 & 68.26 & 64.49 & 83.06 & 83.63 & 84.52 & 59.1 \\
        LAST & 74.85 & 74.08 & 59.15 & 75.17 & \textbf{69.02} & \textbf{66.12} & 83.19 & 83.41 & 85.75 & 61.59 \\
        LAVISH-Att & 75.32 & 75.47 & \textbf{63.39} & 74.37 & 68.37 & 64.94 & 83.72 & \textbf{84.08} & \textbf{86.32} & 61.74 \\ 
        LAST-Att & \textbf{75.44} & \textbf{76.21} & \textbf{60.60} & \textbf{75.23} & 68.91 & 65.60 & \textbf{84.12} & 84.01 & \textbf{86.03} & \textbf{62.52} \\
        \bottomrule
    \end{tabular}%
    }
    \vspace{-0.3cm}
    \label{tab:ablation_balance}
\end{table}

As shown in the table, we observe that ``Swin-AST" achieves nearly on-par results with our ``LAST" baseline, and performs better than LAVISH with a +1.45\% improvement. This suggests the benefits of applying robust pretrained backbones for both the visual and audio branches for AVQA task. Moreover, ``LAVISH-Att", even without using a pretrained audio backbone, surpasses the LAVISH baseline by +2.14\%. However, it falls short by a mere 0.12\% compared to our full model, ``LAST-Att". This confirms our hypothesis that integrating a fine-grained spatial cross-attention module across feature maps of both modalities can improve performance, especially when combined with the existing spatial grounding module. In our experiments, when we add only the cross-modal pixel-wise attention module and excluded the spatial grounding branch, the model underperform from the outset, plateauing at a total accuracy of 69.3\%. We hypothesize that the fine-grained attention module captures low-level feature details, while the spatial grounding module abstracts high-level features. They complement each other to bring the optimal result.

\end{document}